\documentclass[10pt,twocolumn,letterpaper]{article}

\usepackage{cvpr}              
\usepackage{multirow}
\usepackage{cancel}
\usepackage[normalem]{ulem}
\usepackage{algorithm}
\usepackage{tabularx}
\usepackage{algpseudocode}
\usepackage{pifont}
\usepackage[accsupp]{axessibility}
\definecolor{cvprblue}{rgb}{0.21,0.49,0.74}
\usepackage[pagebackref,breaklinks,colorlinks,allcolors=cvprblue]{hyperref}
\usepackage[table]{xcolor}

\def\paperID{} 
\def\confName{CVPR}
\def\confYear{2026}

\title{ARMFlow: AutoRegressive MeanFlow for Online 3D Human Reaction Generation}

\author{
Zichen Geng\textsuperscript{1},
Zeeshan Hayder\textsuperscript{2},
Wei Liu\textsuperscript{1},
Hesheng Wang \textsuperscript{3}\thanks{Corresponding author},
Ajmal Saeed Mian\textsuperscript{1} \\
\textsuperscript{1}The University of Western Australia, Perth, WA, Australia \\
\textsuperscript{2}Commonwealth Scientific and Industrial Research Organisation (CSIRO), Canberra, ACT, Australia \\
\textsuperscript{3}Shanghai Jiao Tong University, Shanghai, China \\
{\tt\small
\begin{tabular}{c}
zen.geng@research.uwa.edu.au \\
\{wei.liu, ajmal.mian\}@uwa.edu.au \\
zeeshan.hayder@data61.csiro.au \\
wanghesheng@sjtu.edu.cn
\end{tabular}
}
}

\begin{document}
\maketitle
\vspace{-10mm}
\begin{abstract}
3D human reaction generation faces three main challenges: (1) high motion fidelity, (2) real-time inference, and (3) autoregressive adaptability for online scenarios. Existing methods fail to meet all three simultaneously. We propose ARMFlow, a MeanFlow-based autoregressive framework that models temporal dependencies between actor and reactor motions. It consists of a causal context encoder and an MLP-based velocity predictor. We introduce Bootstrap Contextual Encoding (BSCE) in training, encoding generated history instead of the ground-truth ones, to alleviate error accumulation in autoregressive generation. We further introduce the offline variant {ReMFlow}, achieving state-of-the-art performance with the fastest inference among offline methods. Our ARMFlow addresses key limitations of online settings by: (1) enhancing semantic alignment via a global contextual encoder; (2) achieving high accuracy and low latency in a single-step inference; and (3) reducing accumulated errors through BSCE. Our single-step online generation surpasses existing online methods on InterHuman and InterX by about 30\% in FID, while matching offline state-of-the-art performance despite using only partial sequence conditions. The official implementation is publicly available at: {\color{blue}\href{https://github.com/ZenGengChin/armflow_official}{https://github.com/ZenGengChin/armflow\_official}}


\end{abstract}    
\vspace{-4mm}
\section{Introduction}
\label{sec:intro}
\vspace{-1mm}

Recent advances in generative modeling have led to remarkable progress in 3D human motion generation, covering a wide spectrum of tasks and methodologies. These include text-guided motion synthesis \cite{a2m, t2mgpt, temos, mdm, mld, guo2024momask, motiondiffuse, attt2m, fgt2m}, human–object interaction \cite{interdiff, hoidiff, chois, ardhoi, cghoi}, scene-conditioned motion generation, and multi-person interaction modeling. Among these, one particularly distinctive task is human reaction generation, which focuses on producing reactive human behaviors in response to other agents or stimuli. This task holds immediate practical potential in human–robot interaction, augmented, and virtual realities.

\begin{figure}
    \centering
    \includegraphics[width=1\linewidth]{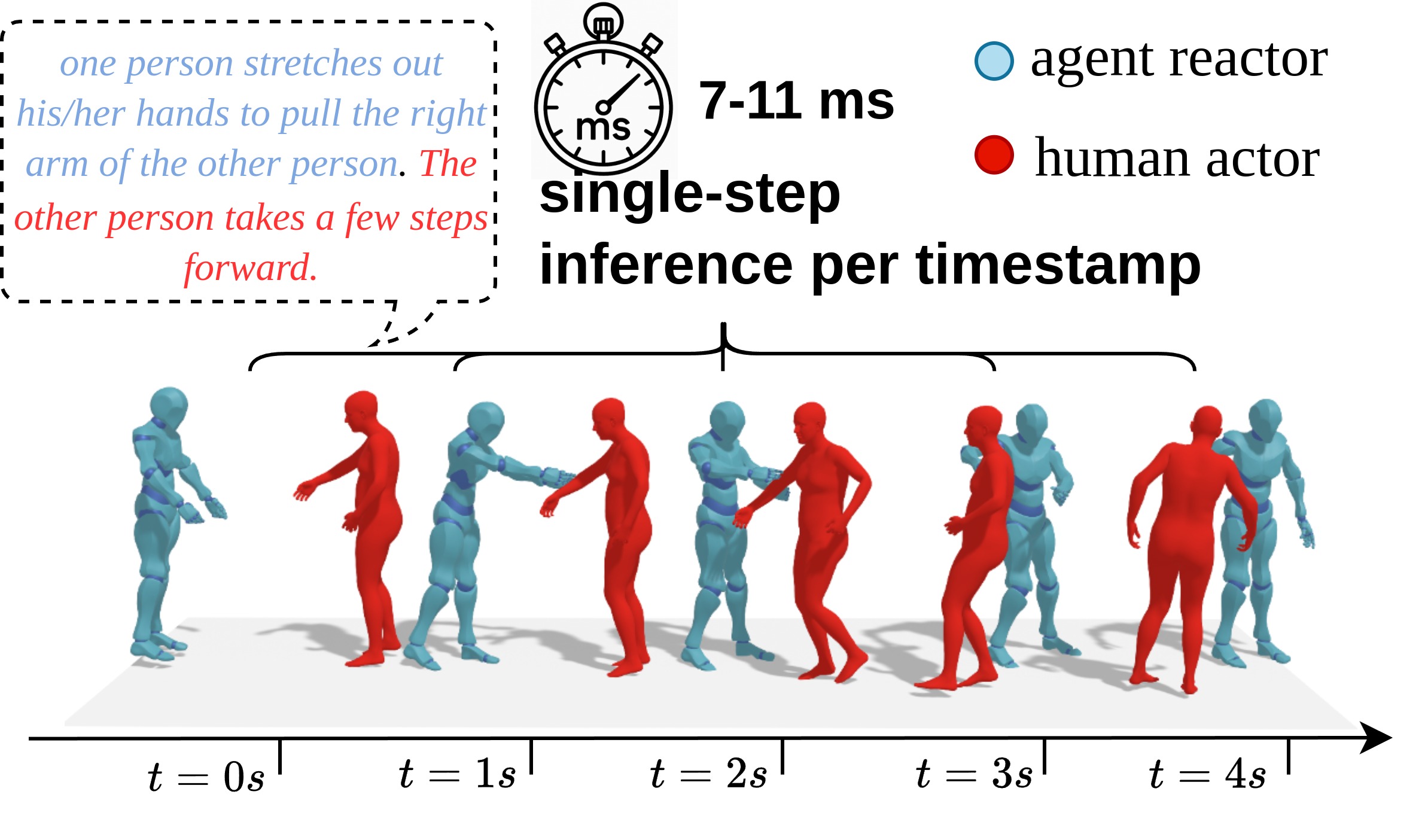}
    \vspace{-7mm}
    \caption{\textcolor{black}{Our method only processes a single inference in each real-time step for online reaction generation, compared to the SOTA methods ReGenNet (35-78 ms), and CAMDM (45 ms). The text description is from the InterX \cite{interx} dataset.}}
    \label{fig:teaser}
    \vspace{-6mm}
\end{figure}

Unlike \textit{offline motion generation} tasks that rely on pre-defined conditions and tolerate second-level latency, \textit{3D human reaction generation} demands real-time responsiveness, where the input condition evolves continuously and unpredictably. In such settings, even minimal computational delays can compromise the system’s reactivity and realism. This imposes stricter requirements on the generation framework, introducing three key challenges:
(1) achieving high inference efficiency to meet real-time constraints;
(2) maintaining high-fidelity motion quality to ensure natural and expressive reactions; and
(3) capturing long-term contextual dependencies to preserve semantic consistency and generate accurate, context-aware motions.

Existing approaches have made attempts to address these challenges, yet fundamental limitations remain. To enable online generation, autoregressive architectures are essential, as they inherently model temporal dependencies between past and future motions. Methods such as CAMDM \cite{camdm} and HumanX \cite{ji2025humanx} adopt autoregressive diffusion models, where a fixed-length historical window serves as the conditioning context for current denoising steps. However, this design faces two key limitations:
(1) the fixed context window hinders scalability and leads to inevitable information loss, causing semantic drift over long sequences; and
(2) although accelerated samplers such as DDIM have been adopted, they still require multiple denoising steps (typically $\geq$8), which is computationally demanding, especially under fine-grained temporal resolutions.

To address the aforementioned three challenges and two key limitations of prior works, we propose ARMFlow, a scalable autoregressive architecture capable of generating high-fidelity human reactions in a \textit{single-step inference}.
Our method is built upon the recently proposed MeanFlow \cite{geng2025mean} paradigm, which enables one-step generation as opposed to multi-step denoising or iterative integration required by diffusion \cite{Diff9D, ardhoi} or traditional flow-based models.
This is the first work that leverages MeanFlow for human motion generation, demonstrating the fastest inference speed while maintaining superior motion realism compared to existing approaches.
To overcome the limitations of fixed-length contextual windows commonly used in prior autoregressive models, we introduce a causal context encoder that encodes the entire motion history with causal masking.
This design prevents the loss of information beyond the context window and preserves global temporal semantics, ensuring coherent long-term motion generation.

Furthermore, we observe that training the model with only clean past motions makes it overly sensitive to noise during autoregressive generation, as it never learns to handle imperfect or accumulated prediction errors in its historical context.
To mitigate this issue, we propose Bootstrap Context Encoding (BSCE), where the model uses predicted motion histories—rather than ground-truth ones during training to construct the contextual conditions for the flow velocity predictor. As training progresses, the predicted motion sequences gradually approximate real trajectories, leading to an adaptive curriculum that naturally reduces context noise over time.
We further increase the number of bootstrap iterations throughout training, effectively introducing controlled noise and enhancing model robustness against accumulated prediction errors. This mechanism accelerates convergence and improves autoregressive stability. Benefiting from MeanFlow's single-step inference, BSCE can efficiently generate augmented history samples without costly iterative denoising, resulting in significantly improved training efficiency.
Together, these components form a self-consistent and efficient autoregressive generation framework based on MeanFlow dynamics.

Beyond the online setting, we further design a general offline variant, Reaction MeanFlow (ReMFlow), which serves as a versatile baseline for offline motion generation.
Compared with existing state-of-the-art (SOTA) offline models, ReMFlow achieves superior generation quality and the fastest inference speed on both InterHuman and InterX benchmarks.
More importantly, our online model ARMFlow not only achieves SOTA performance in real-time settings, notably over 40\% on FID, but also performs on par with, and in some cases surpasses, many offline models that have access to the entire conditioning sequence simultaneously. In summary, our contributions are threefold:
\begin{enumerate}
\item We propose ARMFlow, a flexible and scalable method consisting of a context encoder with an MLP velocity predictor that captures global semantic alignment in an autoregressive manner. 
\item We extend the MeanFlow paradigm to the domain of 3D reaction generation, introducing a unified framework for both online (ARMFlow) and offline (ReMFlow) settings that achieves single-step, high-fidelity motion synthesis with real-time performance.

\item We propose a Bootstrap Context Encoding (BSCE) mechanism that effectively mitigates error accumulation in autoregressive generation, speeds up the model's convergence, and enhances inference robustness.
\end{enumerate}






\section{Related Works}
\label{sec:relatedworks}

\paragraph{Reaction Generation:} 3D human motion generation is currently an active area of research, encompassing tasks such as Text-to-Motion \cite{a2m, temos, mdm, mld, ACTOR, t2mgpt, guo2024momask, attt2m,fgt2m}, Human-Object Interaction \cite{hoidiff, cghoi, chois, ardhoi, interdiff, physhoi}, to Human-Human Interaction \cite{intergen, in2in, wang2025timotion, intermask} generation. However, Action-Reaction Generation is still an understudied area. Tab.~\ref{tab:relatedworks} lists current SOTA models for reaction generation.

\begin{table}[h]
\centering
\renewcommand{\arraystretch}{0.95}  
\caption{Current reaction generation models. AR: Autoregression.}
\vspace{-2mm}
\resizebox{\linewidth}{!}{
\begin{tabular}{l c c c c}
\toprule
{Model} & {Online} & {Real-time} & {Long-context} & {AR} \\
\midrule
InterMask  \cite{intermask} & \ding{55} & \ding{55} & \ding{51} & \ding{55} \\
InterGen  \cite{intergen}  & \ding{55} & \ding{55} & \ding{51} & \ding{55} \\
MARRS \cite{wang2025marrs}& \ding{55} & \ding{51} & \ding{51} & \ding{55} \\
CAMDM   \cite{camdm}    & \ding{51} & \ding{51} & \ding{55} & \ding{51} \\
ReGenNet \cite{xu2024regennet}    & \ding{51} & \ding{51} & \ding{55} & \ding{55} \\
HumanX \cite{ji2025humanx}      & \ding{51} & \ding{51} & \ding{55} & \ding{51} \\
Ours & \ding{51} & \ding{51} & \ding{51} & \ding{51} \\
\bottomrule
\end{tabular}
}
\label{tab:relatedworks}
\vspace{-3mm}
\end{table}

\begin{figure*}[h]
    \centering
    \includegraphics[width=1\linewidth]{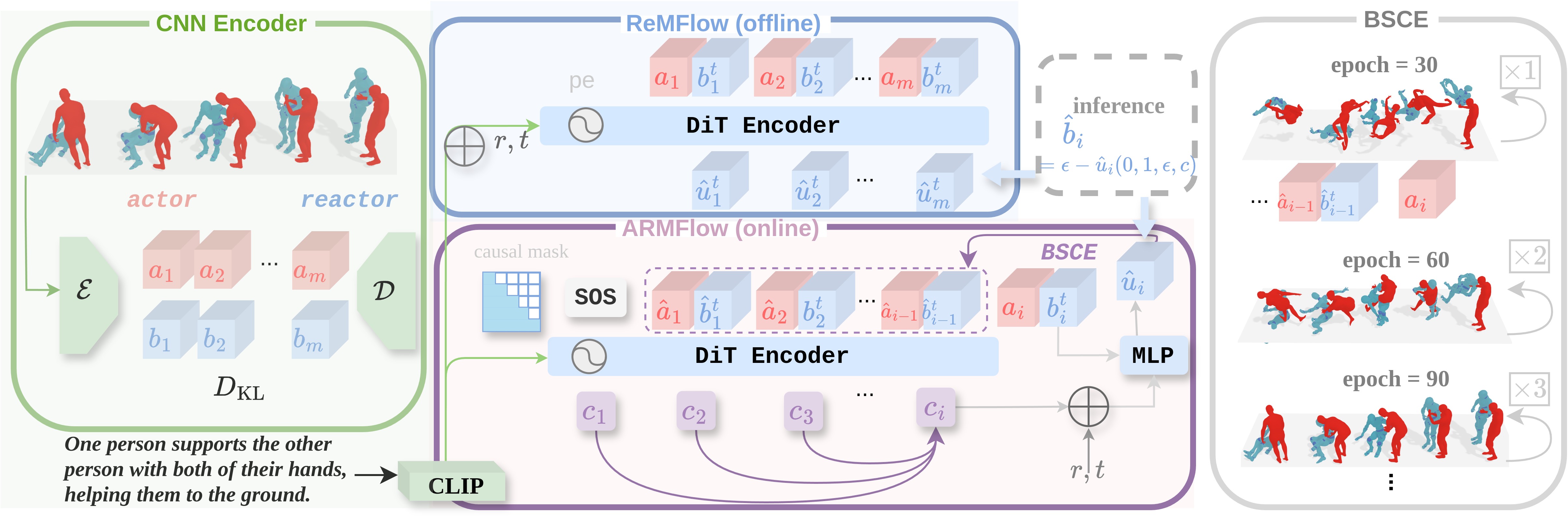}
    \vspace{-5mm}
    \caption{\textcolor{black}{Overview of the proposed architecture for online and offline reaction generation. The framework consists of a CNN-based encoder to learn a compact latent space for the actor and the reactor. The ReMFlow is for offline generation based on the DiT architecture, and ARMFlow is the autoregressive online model consisting of a DiT context encoder and an MLP velocity predictor. A BSCE strategy is employed during online training progressively to reduce accumulated error in the autoregression.}}
    \vspace{-3mm}
    \label{fig:armflow}
\end{figure*}

\citet{interformer} first proposed Interformer, a Transformer-based \cite{vaswani2017attention} reaction synthesis model, but due to the use of traditional autoencoders to predict current actions, its prediction accuracy is less than ideal. \citet{xu2024regennet} introduced the first online Transformer-decoder-based diffusion model, aiming to achieve both efficiency and accuracy in diffusion via DDIM \cite{ddim}. Although this method performs online generation, it inherently lacks autoregression (AR) (i.e., generating the current action based on previously generated content), leading to temporal inconsistency. \citet{ji2025humanx} addressed the autoregressive online generation issue by conditioning on a fixed window of previous context to guide the generation of the next timestep’s action. However, a fixed-window design imposes significant scalability constraints: it cannot be adapted to finer temporal resolutions, and approaches such as \cite{camdm} are unable to perform inference from scratch without any prior context. Moreover, when processing long sequences, the limited receptive field fails to capture distant dependencies, leading to cumulative error and a lack of global contextual understanding. R2R \cite{cen2025ready_to_react} aims to solve this by introducing full-history encoding, but its 50-step inference limits its efficiency. 

\paragraph{Autoregressive generative models} have been extensively explored, particularly for image generation. Some approaches tokenize continuous representations into discrete token spaces in an LLM-style manner \cite{tian2024var, lfqvae, t2mgpt, maskgit}, while others adopt diffusion-based methods \cite{ardiff, ARDM, armd, aamdm} in an autoregressive fashion. Another line of research employs autoregressive flow-based architectures \cite{tokenflow, nextstep}.
Although autoregressive models are capable of capturing strong dependencies within data distributions, a core challenge lies in mitigating error accumulation that naturally occurs during sequential inference. T2M-GPT \cite{t2mgpt} addresses this issue by introducing random token corruption during training, forcing the model to correct corrupted past contexts. Subsequently, \citet{tian2024var} proposed a hierarchical data representation to alleviate cumulative errors. However, these techniques are not directly applicable to conventional autoregressive models—especially diffusion-based architectures with fixed contextual windows \cite{ARDM, armd, aamdm}.
To overcome this limitation, \citet{ardiff} proposed a general framework combining an autoregressive context encoder with a lightweight MLP denoiser, leveraging MAE-style random masking \cite{MAE} to reduce error propagation along the autoregressive direction. Nevertheless, such random masking schemes are less suitable for specially ordered or real-time generation tasks, where temporal consistency is essential. As a result, enhancing the robustness of real-time autoregressive models remains unsolved.


\section{{Preliminary}}

MeanFlow \cite{geng2025mean} is a recently proposed generative paradigm that aims to produce high-fidelity samples with only a single inference step. Unlike traditional Flow Matching (FM) \cite{flowmatching,rectifiedflow} approaches, which learn the instantaneous velocity field by estimating the continuous trajectory between data and noise distributions through numerous integration steps, MeanFlow instead models the mean velocity over the entire trajectory. Specifically, instead of computing the instantaneous flow $\mathbf{v}(x, t)$ at each time step, MeanFlow evaluates the average transport velocity between any two points $r$ and $t$ in the field. This simplification allows the model to generate samples by a single-step integration of the mean field from $r=0$ (noise) to $t=1$ (data), substantially reducing inference time while maintaining strong generative quality.
Our method builds upon {MeanFlow}, enabling efficient reaction generation with single-step inference. Given an instantaneous velocity field $v(z_\tau,\tau)$, the average velocity between timesteps $r$ and $t$ is defined as:
\vspace{-3mm}
\begin{equation}
u(z_t, r, t) = \frac{1}{t-r}\int_{r}^{t} v(z_\tau, \tau)\, d\tau,
\end{equation}
which captures the cumulative dynamics over the interval $[r,t]$ and provides a smooth approximation of trajectory evolution. Differentiating both sides using the Leibniz rule yields the training objective:
\begin{align}
\mathcal{L}(\theta) &= \mathbb{E}\big\|u_{\theta}(z_t, r, t) - \mathrm{sg}(u_{\text{tgt}})\big\|_2^2, \\
u_{\text{tgt}} &= v(z_t, t) - (t-r)\big( v(z_t,t)\,\partial_z u_{\theta} + \partial_t u_{\theta} \big),
\end{align}
where $\mathrm{sg}(\cdot)$ denotes the stop-gradient operator to stabilize optimization. The Jacobian--vector products (JVPs) for $\partial_z u_{\theta}$ and $\partial_t u_{\theta}$ are computed via automatic differentiation, avoiding explicit Jacobian construction and reducing memory overhead while maintaining accurate gradient signals.
Unlike conventional flow-based methods that uniformly sample timestep pairs, we adopt a \emph{biased sampling strategy} where a proportion of pairs satisfy $r = t$. This allocation strengthens the learning of the instantaneous velocity $v(z_t,t)$, improves convergence, and enhances fidelity for high-frequency motion components, which are critical for reactive behaviors.
A key advantage of MeanFlow is {single-step inference}, eliminating iterative refinement,
which drastically reduces inference cost compared to multi-step diffusion or flow-based models, making it suitable for large-scale generation.

To improve controllability without additional inference overhead, \citet{geng2025mean} incorporate \emph{classifier-free guidance} (CFG) during training rather than post-hoc blending by constructing a Ground-Truth Field. The modified objective becomes:
\vspace{-3mm}

\begin{align}
\mathcal{L}(\theta) &= \mathbb{E}\big\| u_{\theta}^{\text{cfg}}(z_t, r, t \mid c) - \mathrm{sg}(u_{\text{tgt}}) \big\|_2^2, \\
u_{\text{tgt}} &= \tilde{v}_t - (t - r)\big( \tilde{v}_t \,\partial_z u_{\theta}^{\text{cfg}} + \partial_t u_{\theta}^{\text{cfg}} \big), \\
\tilde{v}_t &= \omega v_t + (1 - \omega) u_{\theta}^{\text{cfg}}(z_t, t, t),
\end{align}

\vspace{-2mm}

\noindent where $c$ denotes the conditioning signal, and $\omega > 1$ controls the blend between conditional and unconditional velocity fields. Training with CFG not only enhances the model's robustness and fidelity, but also reduces the extra calculation in inference for unconditional output. 

\section{Method}

Given a text description $text$ and an actor motion sequence $x_a \in \mathbb{R}^{T,D}$, where $T$ denotes the number of frames and $D$ the pose dimension, our goal is to synthesize a realistic reactor motion $x_b \in \mathbb{R}^{T,D}$. The generation can occur in two modes: \emph{sequential (online)}, where frames are predicted progressively, or \emph{offline}, where the entire sequence is generated in a single pass. To achieve this, we first compress actor and reactor motions into a shared latent space using the CNN-VAE described in Section~\ref{subsec:cnn}, which benefits both offline and online reaction generation. Building on this representation, we introduce \textbf{ReMFlow} (Section~\ref{subsec:meanflow}) for offline generation and \textbf{ARMFlow} (Section~\ref{subsec:armflow}) for online generation, both leveraging the MeanFlow framework and a DiT-based backbone~\cite{dit}. Finally, we propose the \emph{Bootstrap Context Encoding} (BCSE), which is inherently suited for ARMFlow and mitigates error accumulation during autoregression. 

\subsection{CNN-VAE for Motion Compression}\label{subsec:cnn}
To improve generation efficiency and robustness, we adopt a 1D-CNN VAE following T2M-GPT \cite{t2mgpt} to compress actor–reactor motion sequences $\mathbf{x}=\{\mathbf{x}_a,\mathbf{x}_b\}$ into temporal latent representations $\{\mathbf{a},\mathbf{b}\}$. This serves to (1) reduce motion dimensionality for efficient computation and (2) provide a structured latent space for autoregressive modeling.

Unlike canonicalized actor motion, reactor motion is highly dynamic and lacks standardization, making discrete quantization prone to accuracy loss. We therefore use continuous latent tokens with KL-divergence regularization instead of discrete codes. This embeds motion into a compact yet expressive latent space, accelerating convergence and improving reconstruction. The causal 1D-CNN preserves temporal dependencies, enabling efficient disentanglement of latent tokens in online generation and enhancing flexibility in sequential contexts. The VAE objective is:
\begin{align}
\mathcal{L}_{\text{VAE}} &= \mathbb{E}_{q(z|x)}[\log p(x|z)] - \mathrm{KL}(q(z|x)\|p(z)),
\end{align}
\noindent where $q(z|x)$ is the encoder, $p(x|z)$ the decoder, and $p(z)$ the Gaussian prior. To ensure high-fidelity reconstruction, we employ inverse kinematic (IK) loss for joint positions and velocity loss for motion smoothness.

\subsection{ReMFlow for Offline Reaction Generation}\label{subsec:meanflow}

We implement \textbf{ReMFlow} for offline generation using a \textbf{DiT}-based encoder with multimodal conditioning. Text prompts are encoded via a CLIP text encoder \cite{CLIP} and fused with timestep embeddings to form a global condition vector, which is injected into DiT through adaptive layer normalization  (AdaLN) to modulate intermediate activations. Actor tokens $a_i$ are concatenated with interpolated reactor tokens $b_i^{t}$, encoded from a CNN-based VAE to capture local appearance and dynamics of the reactor. These tokens are aligned along the last dimension, summed with positional encodings, and processed by DiT to predict average velocities $\mathbf{u}=\{u_i\}_{i=1}^m$ over tokenized spatiotemporal patches. To enable unconditional learning and support CFG, a proportion of text and actor tokens are replaced with null tokens $\varnothing$ during training, which regularizes the model and stabilizes optimization under varying conditioning strengths.

This design allows ReMFlow to jointly model actor dynamics and reactor responses under flexible conditioning while maintaining computational efficiency. Compared to iterative diffusion models, our approach achieves significant speedup and scalability due to single-step inference and average-velocity prediction, making it practical for large-scale offline reaction generation tasks without sacrificing fidelity or semantic alignment.

\subsection{ARMFlow for Online Generation}\label{subsec:armflow}

\textbf{ARMFlow} overcomes the limitations of fixed-window architectures by enabling encoding from scratch and retaining all past information, thereby preserving global semantics. As illustrated in Fig.~\ref{fig:armflow}, the architecture follows the spirit of MAR~\cite{ardiff}, comprising a DiT-based context encoder and a lightweight MLP velocity predictor. The former encodes historical motion and text semantics and supplies them as partial conditions to the MLP predictor.

During training, we concatenate actor and reactor history tokens $(a_i, b_i)$ along the last dimension and prepend a learnable start-of-sequence token $\langle \text{sos} \rangle$ to ensure that inference remains feasible at $t{=}0$. The concatenated sequence is fed into the DiT backbone, while text conditioning is injected through normalization layers (via infusion), and a causal mask is applied to enforce forward-only temporal dependencies for autoregressive learning. The encoded historical context $c_i$ is then combined with the upsampled timesteps $(r, t)$ and used as conditions for an AdaLN-modulated MLP. At the current step, the interpolated reactor sample $b^{r,t}_i$ from the velocity field is passed to the MLP velocity predictor, which outputs the average velocity $\hat{u}_{r,t}$. Unlike offline generation, the conditioning input here includes not only the current actor but also the accumulated history; consequently, when performing classifier-free guidance (CFG), we augment the null token with a \emph{null history} to match the online conditioning interface.

At inference time, the start token $\langle \text{sos} \rangle$ is first encoded as the initial history. A Gaussian noise token $\epsilon_1$ is then paired with the current actor token $a_1$ and fed into the MLP with $(r{=}0,\,t{=}1)$ to predict the average velocity $\hat{u}_1$. The current reactor token is obtained by $b_1 = \epsilon_1 - \hat{u}_1$. The predicted reactor token is cached together with the corresponding actor token to update the history, and the process is iterated autoregressively until actor tokens end.



\subsection{Bootstrap Contextual Encoding}\label{subsec:method_bsce}

Drift over long sequences is an inherent limitation of autoregressive models. HumanX \cite{ji2025humanx} addresses this through a history-rollout training strategy, where the ground-truth (GT) history condition is gradually replaced by the model’s generated history. However, this approach has three main drawbacks. First, to stabilize training, HumanX gradually subtracts the GT history from the generated ones, slowing the convergence. Second, as the model converges, the generated history becomes increasingly similar to the GT, leading to insufficient self-augmentation. Third, HumanX only replaces the reactor history while keeping the actor unchanged, which lead to overfitting to the actor’s motion.

To address these limitations, we propose Bootstrap Contextual Encoding (BSCE). As detailed in Algorithm 1, 
BSCE replaces both actor and reactor histories with generated samples from the very beginning of training. As the model progressively aligns its outputs with the GT, the number of autoregressive iterations is gradually increased on schedule. This amplified accumulated error introduces additional noise, which in turn enhances the model’s robustness and generalization.
Moreover, since HumanX relies on a multi-step diffusion model, its rollout is computationally expensive during the training phase. In contrast, our MeanFlow-based generator performs single-step inference, providing substantial efficiency gains and further highlighting BSCE's natural compatibility with MeanFlow.


\begin{algorithm}[t]
\caption{Bootstrap Contextual Encoding (BSCE)}
\label{alg:bsce_simplified}
\begin{algorithmic}[1]
\Procedure{BSCE}{$G_\theta, \mathcal{Y}, \mathcal{X}, \mathcal{C}, \mathcal{S}_t, K_{\max}, I_{\max}$}
    \State $K \gets 1$
    \For{iteration $= 1$ to $I_{\max}$}
        \State Sample $(x_a, x_b, c)$ from $\mathcal{Y}, \mathcal{X}, \mathcal{C}$
        \State Initialize context buffer $\mathcal{Z} \gets \{\langle \text{sos} \rangle\}$
        \State $(\{a_i\}, \{b_i^{\text{gt}}\}) \gets \textsc{Tokenize}(x_a, x_b)$
        \For{$i = 1$ to $K$}
            \State $c_i \gets \textsc{EncodeContext}(\mathcal{Z}, c)$
            \State Sample $(r, t) \sim \mathcal{S}_t$, $\epsilon_i \sim \mathcal{N}(0, I)$
            \State $\hat{u}_{r,t} \gets G_\theta(\epsilon_i, a_i \text{ or } b_i, r, t, c_i)$ 
            \State $a_i \text{ or } b_i \gets \epsilon_i - \hat{u}_{r,t}$
            \State Append $(a_i, b_i)$ to $\mathcal{Z}$
            \State $\mathcal{L}_i \gets \textsc{Loss}(\hat{u}_{r,t})$ 
        \EndFor
        \State Update $\theta \gets \theta - \eta \nabla_\theta \left(\frac{1}{K} \sum_i \mathcal{L}_i \right)$
        \State $K \gets \textsc{Scheduled}(iteration, K_{\max})$
    \EndFor
\EndProcedure
\end{algorithmic}
\end{algorithm}
\vspace{-3mm}

\begin{table*}[t!]
\caption{Comparison of online methods on InterHuman and InterX datasets.}
\label{tab:onine_table}
\centering
\resizebox{\linewidth}{!}{%
\renewcommand{\arraystretch}{0.8}
\begin{tabular}{clcccccccc}
\toprule[1.5pt]
Dataset & Model & FID $\downarrow$ & R-Prec@1 $\uparrow$ & R-Prec@2 $\uparrow$ & R-Prec@3 $\uparrow$ & MM Dist $\downarrow$ & Diversity $\rightarrow$ & MModality $\uparrow$ \\
\midrule
\multirow{5}{*}{InterHuman} 
& Ground Truth    & $0.273^{\pm .007}$ & $0.452^{\pm .008}$ & $0.610^{\pm .009}$ & $0.701^{\pm .008}$ & $3.755^{\pm .008}$ & $7.948^{\pm .064}$ & - \\
& InterFormer \cite{interformer} & $4.871^{\pm .049}$ & ${0.302}^{\pm .004}$ & ${0.457}^{\pm .004}$ & ${0.542}^{\pm .005}$ & ${3.845}^{\pm .001}$ & $7.482^{\pm.045}$ & ${0.254}^{\pm.029}$ \\
& CAMDM \cite{camdm} & ${4.000}^{\pm .046}$ & ${0.335}^{\pm .005}$ & ${0.492}^{\pm .005}$ & ${0.587}^{\pm .005}$ & ${3.828}^{\pm .001}$ & ${7.547}^{\pm.025}$ & $\textbf{1.581}^{\pm.026}$ \\
& ReGenNet \cite{xu2024regennet} & ${4.176}^{\pm .085}$ & ${0.355}^{\pm .005}$ & ${0.508}^{\pm .005}$ & ${0.600}^{\pm .004}$ & ${3.817}^{\pm .001}$ & ${7.480}^{\pm .033}$ & ${0.442}^{\pm.012}$ \\
& R2R \cite{cen2025ready_to_react} & $\underline{2.795}^{\pm .062}$ & $\underline{0.431}^{\pm .005}$ & $\underline{0.591}^{\pm .004}$ & $\underline{0.674}^{\pm .004}$ & $\underline{3.793}^{\pm .002}$ & $\underline{7.693}^{\pm .028}$ & $\underline{0.517}^{\pm.013}$ \\
\rowcolor{gray!15}
& ARMFlow (\textbf{Ours}) & $\textbf{2.178}^{\pm .054}$ & $\textbf{0.441}^{\pm .005}$ & $\textbf{0.605}^{\pm .005}$ & $\textbf{0.699}^{\pm .005}$ & $\textbf{3.783}^{\pm .002}$ & $\textbf{7.745}^{\pm .024}$ & ${0.369}^{\pm.008}$ \\

\midrule
\multirow{5}{*}{InterX}
& Ground Truth    & $0.002^{\pm .000}$ & $0.435^{\pm .005}$ & $0.628^{\pm .004}$ & $0.736^{\pm .004}$ & $3.574^{\pm .013}$ & $8.947^{\pm .078}$ & - \\
& InterFormer \cite{interformer} & ${0.304}^{\pm .009}$ & ${0.301}^{\pm .003}$ & ${0.469}^{\pm .003}$ & ${0.571}^{\pm .002}$ & ${4.604}^{\pm .009}$ & ${8.579}^{\pm .061}$ & ${0.289}^{\pm .009}$ \\
& CAMDM \cite{camdm} & ${0.429}^{\pm .011}$ & ${0.312}^{\pm .004}$ & ${0.480}^{\pm .003}$ & ${0.587}^{\pm .003}$ & ${4.468}^{\pm .020}$ & ${8.467}^{\pm .072}$ & $\textbf{1.460}^{\pm .027}$ \\
& ReGenNet \cite{xu2024regennet} & ${0.071}^{\pm .003}$ & ${0.402}^{\pm .005}$ & ${0.584}^{\pm .004}$ & ${0.690}^{\pm .004}$ & ${3.843}^{\pm .011}$ & $\underline{9.011}^{\pm .053}$ & ${0.738}^{\pm .021}$ \\
& R2R \cite{cen2025ready_to_react} & $\underline{0.063}^{\pm .003}$ & $\underline{0.412}^{\pm .005}$ & $\underline{0.598}^{\pm .005}$ & $\underline{0.704}^{\pm .004}$ & $\underline{3.745}^{\pm .011}$ & ${8.873}^{\pm .055}$ & ${1.074}^{\pm.019}$ \\
\rowcolor{gray!15}
& ARMFlow (\textbf{Ours}) & $\textbf{0.042}^{\pm .003}$ & $\textbf{0.420}^{\pm .004}$ & $\textbf{0.606}^{\pm .004}$ & $\textbf{0.711}^{\pm .004}$ & $\textbf{3.728}^{\pm .012}$ & $\textbf{8.939}^{\pm .071}$ & $\underline{1.203}^{\pm .029}$ \\

\bottomrule[1.5pt]
\end{tabular}%
}
\vspace{-2mm}
\end{table*}

\section{Experiments}

\paragraph{Datasets.}
We evaluate our approach on two widely adopted benchmarks for text-conditioned human interaction synthesis: \textit{InterHuman}~\cite{intergen} and \textit{InterX}~\cite{interx}. \textit{InterHuman} comprises 7,779 interaction sequences, while \textit{InterX} provides 11,388 sequences, each annotated with 3 textual descriptions. Compared with the other noisy and small datasets like NTU-120 and CHI3D\cite{ntu_rgbd120, chi3d}, these two datasets are of better motion qualities and have solid and public metrics for evaluations for fair comparisons. 

The \textit{InterHuman} dataset is built upon the AMASS~\cite{amass} skeleton, which includes 22 joints with the root joint. Each joint is represented following the HumanML3D~\cite{a2m} convention as $\{ \mathbf{p}_g, \mathbf{v}_g, \mathbf{r}_{6d} \}$, where $\mathbf{p}_g \in \mathbb{R}^3$ denotes global position, $\mathbf{v}_g \in \mathbb{R}^3$ global velocity, and $\mathbf{r}_{6d} \in \mathbb{R}^6$ the local 6D rotation. This results in a motion tensor $\mathbf{m}_p \in \mathbb{R}^{N \times 22 \times 12}$, accompanied by a binary foot-contact indicator $fc \in \mathbb{R}^2$.

In contrast, \textit{InterX} adopts the SMPL-X~\cite{smplx} skeleton, which has 55 articulated joints covering the body, hands, and facial regions, along with the root orientation. Each joint rotation and root orientation is encoded as $\mathbf{r}_{6d}$, while the root translation and rotation are represented by $\mathbf{t}_r$ and $\mathbf{r}_r$, respectively, forming $\mathbf{m}_p \in \mathbb{R}^{N \times 56 \times 6}$. 

\vspace{-3mm}
\paragraph{Evaluation Metrics.} For reaction generation, we mainly focus on fidelity and semantic correspondence. To comprehensively evaluate the performance of our model, we adopt a suite of feature-space metrics by \citet{intergen}. To assess the realism and fidelity of generated interactions, we compute the Fréchet Inception Distance (FID) and between the feature distributions of generated and ground-truth motions and their Diversity. To evaluate the semantic alignment between text prompts and generated motions, we use R-Precision and Multimodal Distance (MMDist), which measure the consistency between text input and generated motions. To assess generative quality beyond accuracy, we report Multimodality, quantifying the model's ability to produce multiple plausible motions for the same text prompts.

\begin{table*}[h!]
\caption{Comparison of offline methods on InterHuman and InterX datasets.}
\label{tab:offline_table}
\centering
\resizebox{\linewidth}{!}{%
\renewcommand{\arraystretch}{1.}
\begin{tabular}{clcccccccc}
\toprule[1.5pt]
Dataset & Model & FID $\downarrow$ & R-Prec@1 $\uparrow$ & R-Prec@2 $\uparrow$ & R-Prec@3 $\uparrow$ & MM Dist $\downarrow$ & Diversity $\rightarrow$ & MModality $\uparrow$ \\
\midrule
\multirow{6}{*}{InterHuman} 
& Ground Truth    & $0.273^{\pm .007}$ & $0.452^{\pm .008}$ & $0.610^{\pm .009}$ & $0.701^{\pm .008}$ & $3.755^{\pm .008}$ & $7.948^{\pm .064}$ & - \\
& InterGen \cite{intergen} & $9.183^{\pm .174}$ & $0.325^{\pm .004}$ & $0.467^{\pm .004}$ & $0.546^{\pm .005}$ & $3.859^{\pm .001}$ & $7.305^{\pm .047}$ & $1.270^{\pm .023}$ \\
& in2IN \cite{in2in} & $7.913^{\pm .251}$ & $0.362^{\pm .004}$ & $0.504^{\pm .008}$ & $0.589^{\pm .007}$ & $3.832^{\pm .002}$ & $7.709^{\pm .040}$ & $1.165^{\pm .034}$ \\
& MLD$^*$ \cite{mld} & $3.588^{\pm .076}$ & $0.405^{\pm .005}$ & $0.561^{\pm .007}$ & $0.649^{\pm .006}$ & $3.798^{\pm .002}$ & $7.663^{\pm .040}$ & $1.124^{\pm .048}$ \\
& ReGenNet$^*$ \cite{xu2024regennet} & $\underline{2.930}^{\pm .052}$ & ${0.362}^{\pm .005}$ & ${0.513}^{\pm .005}$ & ${0.605}^{\pm .004}$ & $3.815^{\pm .001}$ & ${7.582}^{\pm .064}$  & $\textbf{1.737}^{\pm.020}$ \\
& InterMask \cite{intermask} & ${3.453}^{\pm .061}$ & $\underline{0.451}^{\pm .007}$ & $\underline{0.610}^{\pm .006}$ & $\underline{0.701}^{\pm .005}$ & $\underline{3.782}^{\pm .002}$ & $\underline{7.710}^{\pm .046}$  & $\underline{1.361}^{\pm.032}$ \\
\rowcolor{gray!15}
& ReMFlow (\textbf{Ours}) & $\textbf{2.433}^{\pm .042}$ & $\textbf{0.452}^{\pm .006}$ & $\textbf{0.618}^{\pm .005}$ & $\textbf{0.708}^{\pm .005}$ & $\textbf{3.778}^{\pm .001}$ & $\textbf{7.714}^{\pm .029}$  & ${0.685}^{\pm.018}$ \\
\midrule
\multirow{5}{*}{InterX}
& Ground Truth & $0.002^{\pm .000}$ & $0.435^{\pm .005}$ & $0.628^{\pm .004}$ & $0.736^{\pm .004}$ & $3.574^{\pm .013}$ & $8.947^{\pm .078}$ & - \\
& InterGen \cite{intergen} & $0.238^{\pm .038}$ & $0.352^{\pm .004}$ & $0.542^{\pm .005}$ & $0.643^{\pm .004}$ & $4.212^{\pm .029}$ & $8.773^{\pm .067}$ & $1.552^{\pm .029}$ \\
& MLD$^*$ \cite{mld} & $0.148^{\pm .020}$ & $0.414^{\pm .004}$ & $0.607^{\pm .006}$ & $0.712^{\pm .004}$ & $3.655^{\pm .020}$ & $8.893^{\pm .053}$ & $\underline{1.875}^{\pm.088}$ \\
& ReGenNet \cite{xu2024regennet} & $\underline{0.093}^{\pm .022}$ & $0.407^{\pm .004}$ & $0.589^{\pm .004}$ & $0.705^{\pm .003}$ & $3.762^{\pm .015}$ & $8.841^{\pm .072}$ & $\textbf{1.937}^{\pm .069}$ \\
& InterMask \cite{intermask} & $0.399^{\pm .013}$ & $\underline{0.429}^{\pm .005}$ & $\underline{0.622}^{\pm .005}$ & $\underline{0.731}^{\pm .005}$ & ${3.584}^{\pm .017}$ & ${8.911}^{\pm .057}$ & $0.859^{\pm.033}$ \\
\rowcolor{gray!15}
& ReMFlow \textbf{(Ours)} & $\textbf{0.058}^{\pm .005}$ & $\textbf{0.440}^{\pm .004}$ & $\textbf{0.636}^{\pm .004}$ & $\textbf{0.743}^{\pm .003}$ & $\textbf{3.570}^{\pm .013}$ & $\textbf{8.948}^{\pm .068}$ & $1.607^{\pm .039}$ \\
\bottomrule[1.5pt]
\end{tabular}%
}
\vspace{-2mm}
\end{table*}

\vspace{-3mm}
\paragraph{Baselines.} We divide our evaluation into two parts. For offline generation, we compare our approach with several state-of-the-art methods, including In2IN, which first synthesizes actor--reactor motions with an MDM prior and subsequently refines them; ReGenNet, a diffusion model with a transformer-decoder backbone; InterMask, a discrete autoregressive model based on masking; and an extended variant of MLD that performs diffusion in a latent space. For online generation, we benchmark against InterFormer, a conventional transformer-based autoregressive architecture; the online configuration of ReGenNet; CAMDM, which adopts a transformer encoder and conditions on a fixed-length context window; and R2R, an autoregressive diffusion diagram. Notably, these online baselines are not explicitly designed for real-time synthesis. Although ReGenNet can apply causal masking, its decoder architecture only conditions on the actor’s previous frame and cannot exploit the reactor’s past states, which often leads to temporal discontinuities. CAMDM, on the other hand, relies on a short fixed window of 10 frames, limiting long-range contextual modeling and causing pronounced drift in extended sequences, particularly on InterHuman. Furthermore, these designs do not support generation from scratch, rendering inference at $t{=}0$ infeasible. In contrast, our method operates at a finer 4-frame granularity for real-time synthesis, scales more favorably than CAMDM, and does not depend on any initial motion. To ensure a fair comparison, we provide CAMDM with additional support by supplying an initial motion sequence, which compensates for its inability to generate from scratch, while keeping its 10-frame context window as specified by the authors.


\vspace{-3mm}
\paragraph{Implementation Details.}
For the CNN encoder, we use the same architecture for both datasets except for the input dimension. Following T2M-GPT, the encoder has a hidden size of 256, two residual convolutional downsampling blocks, and three hidden layers per block. The main model adopts a standard DiT backbone for both ARMFlow (online) and ReMFlow (offline), with 512 hidden dimensions, 7 transformer layers, and 8 attention heads with skip connections. The MLP-based velocity predictor in ARMFlow is a lightweight 5-layer network, with identical settings across datasets. For MeanFlow, timesteps are sampled from a logit-normal distribution, with the probability of instantaneous velocity sampling ($r=t$) set to 0.25. In ReMFlow, classifier-free guidance (CFG) is set to $\omega=1.8$ for InterHuman and $\omega=2.0$ for InterX. For ARMFlow, we use $\omega=1.8$ on InterHuman and $\omega=1.2$ on InterX. All models are trained with a batch size of 64. The CNN-VAE is trained for 2000 epochs on InterHuman and 800 on InterX. ReMFlow is trained for 800 and 500 epochs, while ARMFlow converges faster with 500 and 300 epochs, respectively. All experiments use the AdamW optimizer with a learning rate of $1\times10^{-4}$ and are conducted on NVIDIA H100 GPUs in an HPC environment.


\begin{figure*}[h]
    \centering
    \includegraphics[width=1\linewidth]{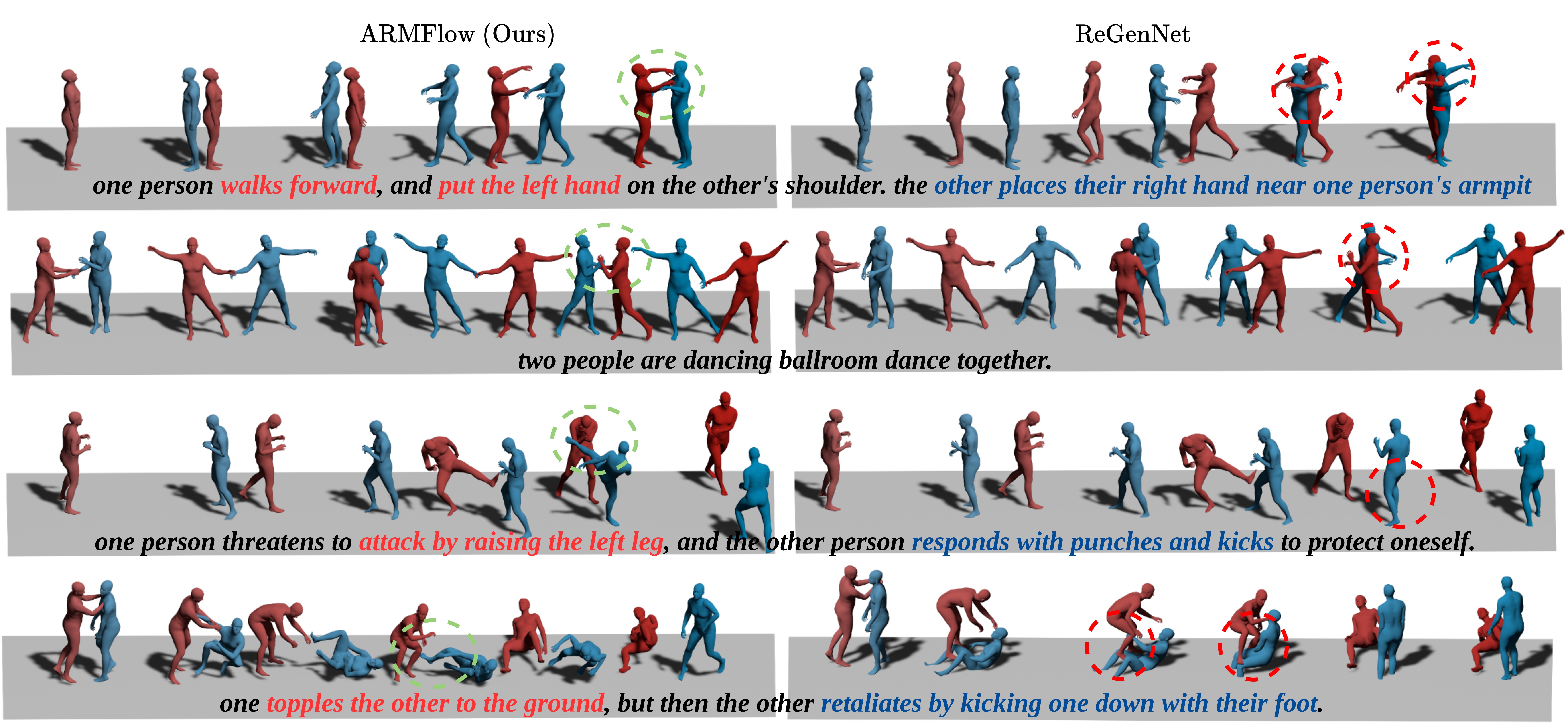}
    \caption{Qualitative comparison with ReGenNet on InterHuman dataset. The problematic interactions are marked with red dashed lines, including penetrations and semantic misalignment, and the correct ones are marked with green.}
    \label{fig:cmpx}
    \vspace{-3mm}
\end{figure*}

\subsection{Quantative Results}


\paragraph{Online Generation.} We first consider the more challenging online generation setting. Tab.~\ref{tab:onine_table} compares our ARMFlow with existing online models, ReGenNet and CAMDM. On the long-sequence InterHuman dataset, both baselines fail to achieve satisfactory results. CAMDM is mainly limited by its short context window, which prevents encoding the full history and leads to severe temporal drift. Although ReGenNet leverages the full actor history, it initializes the reactor input from Gaussian noise at each inference step, preventing adaptive refinement of the reactor representation based on its own history and resulting in temporally inconsistent actions. In contrast, ARMFlow significantly outperforms existing methods in both FID and semantic consistency. Compared with R2R (the second-best in FID), ARMFlow achieves a 28\% relative improvement. Moreover, on the shorter-sequence InterX dataset, our method also achieves state-of-the-art performance. ReGenNet performs relatively well on this dataset, likely because its short-term attention mechanism is more effective for shorter sequences. Notably, online generation does not necessarily degrade performance—in our experiments, ARMFlow even achieves a lower (better) FID in the online setting than in the offline one, while the offline version, benefiting from full actor-conditioned information, yields slightly higher semantic consistency.

\vspace{-3mm}
\paragraph{Offline Generation.} For the offline generation task, we evaluate the generality of ReMFlow and compare it with current SOTA methods in Tab.~\ref{tab:offline_table}.
InterMask, based on discrete random masking, performs well on semantic metrics such as R-precision and MM-Distance but yields a significantly worse FID. This mainly results from its shared VQ-VAE encoder, which over-compresses motion features and degrades reconstruction quality—one reason we avoid discrete representations. ReGenNet operates directly on raw data without compression and thus avoids this loss; however, on the long-sequence InterHuman dataset its self-attention struggles to capture fine-grained semantics, leading to weaker semantic consistency despite a strong FID. MLD lies between these extremes, reflecting a common trend in motion generation: stronger compression improves semantic alignment but often reduces reconstruction fidelity. InterMask achieves high semantic consistency at the cost of efficiency, requiring at least 20 inference steps (0.77 s). In contrast, ReMFlow attains state-of-the-art performance with a single forward pass, greatly improving inference speed while maintaining generation quality. Similar trends appear on the InterX dataset, where shorter sequences allow ReGenNet to achieve relatively strong semantic alignment.




\subsection{Qualitative Comparison}
We present online generation results qualitatively on the InterHuman dataset, which has longer sequences, and the competition is more challenging.
We compare our method with ReGenNet, and visualize the results in Fig.~\ref{fig:cmpx}, which indicate that our approach not only achieves better inter-person contact quality but also demonstrates stronger semantic alignment with the text prompts.

\subsection{Ablation Study}

We validate the effectiveness of our proposed designs and individual modules.
First, for the generative approaches, we kept the same network architecture while replacing the objective function with Diffusion Models and Rectified Flow, respectively, and performed task-specific parameter tuning for them including CFG strength. Experimental results in Tab.~\ref{tab:abl_online_mehtods} and Tab.~\ref{tab:abl_offline_mehtods} 
show that MeanFlow consistently achieves the best FID scores and semantic alignment across both online and offline generation tasks, as well as on both datasets. Moreover, its single-step generation greatly accelerates the training process when integrated with BSCE, demonstrating the broad adaptability of MeanFlow to various generation tasks.
Furthermore, we compare BSCE with two alternative strategies: using ground-truth contextual encoding (GTE) and progressive rollout. BSCE significantly outperforms both approaches, confirming the effectiveness and superiority of our proposed method.


\begin{table}[h]
\caption{Ablation study for online generation on methods.}
\vspace{-2mm}
\label{tab:abl_online_mehtods}
\centering
\resizebox{\linewidth}{!}{%
\renewcommand{\arraystretch}{1.}
\begin{tabular}{clccccccc}
\toprule[1.5pt]
 & Model & R-Prec@3 $\uparrow$ & FID $\downarrow$ & MM Dist $\downarrow$ & Diversity $\rightarrow$ \\
\midrule
\multirow{5}{*}{\rotatebox[origin=c]{90}{InterHuman}} 
& {DDIM 10} & $0.689^{\pm .004}$ & $3.528^{\pm .049}$ & ${3.803}^{\pm .002}$ &  ${7.691}^{\pm .037}$ \\
& {DDIM 50} & $\underline{0.697}^{\pm .005}$ & $3.449^{\pm .062}$ & ${3.794}^{\pm .002}$ & ${7.702}^{\pm .028}$ \\
& {DDPM}    & $0.686^{\pm .004}$ & $3.757^{\pm .068}$ & ${3.806}^{\pm .002}$ & $\textbf{7.803}^{\pm .035}$ \\
& {Rectified Flow 10}  & $0.692^{\pm .004}$ & $\underline{2.449}^{\pm .062}$ & ${3.796}^{\pm .001}$ & ${7.702}^{\pm .028}$ \\
& ARMFlow \textbf{(Ours)} & $\textbf{0.699}^{\pm .005}$ & $\textbf{2.178}^{\pm .054}$ & $\textbf{3.783}^{\pm .002}$ & $\underline{7.745}^{\pm .024}$ \\
\midrule
\multirow{5}{*}{\rotatebox[origin=c]{90}{InterX}} 
& {DDIM 10}        & ${0.692}^{\pm .004}$ & ${0.093}^{\pm .005}$ & ${3.802}^{\pm .014}$ & ${8.870}^{\pm .066}$ \\
& {DDIM 50}        & $\underline{0.705}^{\pm .004}$ & ${0.064}^{\pm .004}$ & $\underline{3.733}^{\pm .013}$ & ${8.895}^{\pm .062}$ \\
& {DDPM}           & ${0.695}^{\pm .004}$ & ${0.106}^{\pm .006}$ & ${3.792}^{\pm .015}$ & $\underline{8.920}^{\pm .088}$ \\
& {Rectified Flow 10} & ${0.698}^{\pm .005}$ & $\underline{0.059}^{\pm .004}$ & ${3.784}^{\pm .011}$ & ${8.913}^{\pm .072}$ \\
& ARMFlow \textbf{(Ours)} & $\textbf{0.711}^{\pm .004}$ & $\textbf{0.042}^{\pm .003}$ & $\textbf{3.728}^{\pm .012}$ & $\textbf{8.939}^{\pm .071}$ \\
\bottomrule[1.5pt]
\end{tabular}%
}
\end{table}

\begin{table}[h]
\caption{Ablation study for offline generation on methods.}
\vspace{-2mm}
\label{tab:abl_offline_mehtods}
\centering
\resizebox{\linewidth}{!}{%
\renewcommand{\arraystretch}{1.}
\begin{tabular}{clccccccc}
\toprule[1.5pt]
 & Model & R-Prec@3 $\uparrow$ & FID $\downarrow$ & MM Dist $\downarrow$ & Diversity $\rightarrow$ \\
\midrule
\multirow{5}{*}{\rotatebox[origin=c]{90}{InterHuman}} 
& {DDIM 10}        & ${0.677}^{\pm .006}$ & ${3.931}^{\pm .050}$ & ${3.800}^{\pm .002}$ & ${7.622}^{\pm .025}$ \\
& {DDIM 50}        & $\underline{0.694}^{\pm .004}$ & ${2.918}^{\pm .068}$ & $\underline{3.789}^{\pm .003}$ & ${7.656}^{\pm .040}$ \\
& {DDPM}           & ${0.681}^{\pm .005}$ & ${3.595}^{\pm .051}$ & ${3.797}^{\pm .002}$ & ${7.663}^{\pm .035}$ \\
& {Rectified Flow 10} & ${0.678}^{\pm .005}$ & $\underline{2.906}^{\pm .044}$ & ${3.801}^{\pm .002}$ & $\textbf{7.775}^{\pm .045}$ \\
& ReMFlow(\textbf{Ours}) & $\textbf{0.708}^{\pm .005}$ & $\textbf{2.433}^{\pm .042}$ & $\textbf{3.778}^{\pm .001}$ & $\underline{7.714}^{\pm .029}$ \\
\midrule
\multirow{5}{*}{\rotatebox[origin=c]{90}{InterX}} 
& {DDIM 10}        & ${0.699}^{\pm .005}$ & ${0.127}^{\pm .036}$ & ${3.805}^{\pm .012}$ & ${8.804}^{\pm .059}$ \\
& {DDIM 50}        & ${0.700}^{\pm .004}$ & $\underline{0.095}^{\pm .076}$ & ${3.749}^{\pm .013}$ & ${8.842}^{\pm .077}$ \\
& {DDPM}           & ${0.671}^{\pm .005}$ & ${3.595}^{\pm .051}$ & ${3.893}^{\pm .012}$ & ${8.857}^{\pm .063}$ \\
& {Rectified Flow 10} & $\underline{0.737}^{\pm .007}$ & ${0.103}^{\pm .015}$ & $\underline{3.645}^{\pm .012}$ & $\underline{8.941}^{\pm .087}$ \\
& ReMFlow(\textbf{Ours})  & $\textbf{0.743}^{\pm .003}$ & $\textbf{0.058}^{\pm .005}$ & $\textbf{3.570}^{\pm .013}$ & $\textbf{8.948}^{\pm .068}$ \\
\bottomrule[1.5pt]
\end{tabular}%
}
\end{table}

\begin{table}[h]
\caption{Ablation on the training strategies for online autoregressive diffusion. GTE stands for ground-truth encoding, Rollout is the strategy used in HumanX\cite{ji2025humanx}, and BSCE is our strategy. }
\label{tab:abl_strategy}
\centering
\resizebox{\linewidth}{!}{%
\renewcommand{\arraystretch}{1.}
\begin{tabular}{clcccccc}
\toprule[1.5pt]
 & Model & R-Prec@3 $\uparrow$ & FID $\downarrow$ & MM Dist $\downarrow$ & Diversity $\rightarrow$ \\
\midrule
\multirow{4}{*}{\rotatebox{90}{InterHuman}} 
& Ground Truth       & $0.701^{\pm .008}$ & $0.273^{\pm .007}$ & $3.755^{\pm .008}$ & $7.948^{\pm .064}$ \\
& {ARMFlow-GTE}      & ${0.602}^{\pm .004}$ & ${5.136}^{\pm .040}$ & ${3.813}^{\pm .002}$ & ${7.728}^{\pm .033}$  \\
& {ARMFlow-Rollout}  & ${0.675}^{\pm .005}$ & ${4.161}^{\pm .055}$ & ${3.798}^{\pm .002}$ & $\textbf{7.802}^{\pm .038}$  \\
& {ARMFlow\textbf{(Ours)}}     & $\textbf{0.699}^{\pm .005}$ & $\textbf{2.178}^{\pm .054}$ & $\textbf{3.783}^{\pm .002}$ & ${7.745}^{\pm .024}$  \\
\midrule
\multirow{4}{*}{\rotatebox{90}{InterX}} 
& Ground Truth       & $0.736^{\pm .004}$ & $0.002^{\pm .000}$ & $3.574^{\pm .013}$ & $8.947^{\pm .078}$ \\
& {ARMFlow-GTE}      & ${0.630}^{\pm .003}$ & ${0.548}^{\pm .007}$ & ${4.667}^{\pm .020}$ & ${8.435}^{\pm .068}$ \\
& {ARMFlow-Rollout}  & ${0.670}^{\pm .004}$ & ${0.192}^{\pm .006}$ & ${4.321}^{\pm .013}$ & ${8.743}^{\pm .082}$ \\

& ARMFlow\textbf{(Ours)}  &$\textbf{0.711}^{\pm .004}$ & $\textbf{0.042}^{\pm .003}$ & $\textbf{3.728}^{\pm .012}$ & $\textbf{8.939}^{\pm .071}$ \\
\bottomrule[1.5pt]
\end{tabular}%
}
\vspace{-6mm}
\end{table}

\section{Conclusion}
We presented ARMFlow, the first MeanFlow-based framework for real-time 3D human reaction generation, achieving single-step inference while maintaining high-fidelity and context-awareness. Our causal context encoder and Bootstrap Context Encoding (BSCE) effectively address long-term dependency modeling and autoregressive error accumulation, enabling robust and efficient online generation. We also introduce ReMFlow as a versatile offline baseline, demonstrating state-of-the-art performance with unprecedented inference speed. Overall, our work establishes a scalable and practical framework for both online and offline human motion generation, paving the way for real-world applications in Human-Robot Interaction, AR, and VR.

\textcolor{black}{Despite these strengths, our method has certain limitations. First, from an engineering perspective, the current implementation does not provide elastic delay handling for the autoregressive small window, which may lead to minor asynchronous behaviors when multiple agents interact. Second, due to the nature of MeanFlow, it does not support post-hoc classifier guidance like diffusion-based models, preventing the use of optimization-based corrections to further refine generated motions. Addressing these limitations represents promising directions for future work.}

\section*{Acknowledgements}

This research was supported by the Australian Research Council (ARC) Discovery Project DP240101926. We gratefully acknowledge this support.

{
    \small
    \bibliographystyle{ieeenat_fullname}
    \bibliography{main}

@String(PAMI = {IEEE Trans. Pattern Anal. Mach. Intell.})

@String(IJCV = {Int. J. Comput. Vis.})

@String(CVPR= {IEEE Conf. Comput. Vis. Pattern Recog.})

@String(ICCV= {Int. Conf. Comput. Vis.})

@String(ECCV= {Eur. Conf. Comput. Vis.})

@String(NIPS= {Adv. Neural Inform. Process. Syst.})

@String(TMM  = {IEEE Trans. Multimedia})

@String(ICLR = {Int. Conf. Learn. Represent.})

@String(AAAI = {AAAI})

@String(PAMI  = {IEEE TPAMI})

@String(IJCV  = {IJCV})

@String(CVPR  = {CVPR})

@String(ICCV  = {ICCV})

@String(ECCV  = {ECCV})

@String(NIPS  = {NeurIPS})

@String(TMM   =	{IEEE TMM})

@String(ICLR  = {ICLR})

@inproceedings{intermask,
        title={InterMask: 3D Human Interaction Generation via Collaborative Masked Modeling},
        author={Muhammad Gohar Javed and Chuan Guo and Li Cheng and Xingyu Li},
        booktitle={The Thirteenth International Conference on Learning Representations (ICLR)},
        year={2025},
        url={https://openreview.net/forum?id=ZAyuwJYN8N}
        }

@article{intergen,
  title={Intergen: Diffusion-based multi-human motion generation under complex interactions},
  author={Liang, Han and Zhang, Wenqian and Li, Wenxuan and Yu, Jingyi and Xu, Lan},
  journal={International Journal of Computer Vision (IJCV)},
  pages={1--21},
  year={2024},
  publisher={Springer}
}

@inproceedings{interx,
  title={Inter-x: Towards versatile human-human interaction analysis},
  author={Xu, Liang and Lv, Xintao and Yan, Yichao and Jin, Xin and Wu, Shuwen and Xu, Congsheng and Liu, Yifan and Zhou, Yizhou and Rao, Fengyun and Sheng, Xingdong and others},
  booktitle={CVPR},
  pages={22260--22271},
  year={2024}
}

@conference{amass,
  title = {{AMASS}: Archive of Motion Capture as Surface Shapes},
  author = {Mahmood, Naureen and Ghorbani, Nima and Troje, Nikolaus F. and Pons-Moll, Gerard and Black, Michael J.},
  booktitle = {International Conference on Computer Vision (ICCV)},
  pages = {5442--5451},
  month = oct,
  year = {2019},
  month_numeric = {10}
}

@inproceedings{smplx,
  title = {Expressive Body Capture: {3D} Hands, Face, and Body from a Single Image},
  author = {Pavlakos, Georgios and Choutas, Vasileios and Ghorbani, Nima and Bolkart, Timo and Osman, Ahmed A. A. and Tzionas, Dimitrios and Black, Michael J.},
  booktitle = {Proceedings IEEE Conf. on Computer Vision and Pattern Recognition (CVPR)},
  pages     = {10975--10985},
  year = {2019}
}

@INPROCEEDINGS{ACTOR,
  title     = {Action-Conditioned 3{D} Human Motion Synthesis with Transformer {VAE}},
  author    = {Petrovich, Mathis and Black, Michael J. and Varol, G{\"u}l},
  booktitle = {International Conference on Computer Vision (ICCV)},
  year      = {2021}
}

@inproceedings{a2m,
  title={Action2motion: Conditioned generation of 3d human motions},
  author={Guo, Chuan and Zuo, Xinxin and Wang, Sen and Zou, Shihao and Sun, Qingyao and Deng, Annan and Gong, Minglun and Cheng, Li},
  booktitle={Proceedings of the 28th ACM International Conference on Multimedia (ACM MM)},
  year={2020}
}

@article{hoidiff,
        title={HOI-Diff: Text-Driven Synthesis of 3D Human-Object Interactions using Diffusion Models},
        author={Peng, Xiaogang and Xie, Yiming and Wu, Zizhao and Jampani, Varun and Sun, Deqing and Jiang, Huaizu},
        journal={arXiv preprint arXiv:2312.06553},
        year={2023}
      }

@inproceedings{cghoi,
  title={Cg-hoi: Contact-guided 3d human-object interaction generation},
  author={Diller, Christian and Dai, Angela},
  booktitle={Proceedings of the IEEE/CVF Conference on Computer Vision and Pattern Recognition (CVPR)},
  pages={19888--19901},
  year={2024}
}

@article{chois,
  title={Controllable human-object interaction synthesis},
  author={Li, Jiaman and Clegg, Alexander and Mottaghi, Roozbeh and Wu, Jiajun and Puig, Xavier and Liu, C Karen},
  journal={arXiv preprint arXiv:2312.03913},
  year={2023}
}

@inproceedings{
mdm,
title={Human Motion Diffusion Model},
author={Guy Tevet and Sigal Raab and Brian Gordon and Yoni Shafir and Daniel Cohen-or and Amit Haim Bermano},
booktitle={The Eleventh International Conference on Learning Representations (ICLR)},
year={2023}

}

@inproceedings{t2mgpt,
            title={T2M-GPT: Generating Human Motion from Textual Descriptions with Discrete Representations},
            author={Zhang, Jianrong and Zhang, Yangsong and Cun, Xiaodong and Huang, Shaoli and Zhang, Yong and Zhao, Hongwei and Lu, Hongtao and Shen, Xi},
            booktitle={Proceedings of the IEEE/CVF Conference on Computer Vision and Pattern Recognition (CVPR)},
            year={2023},
          }

@inproceedings{mld,
  title={Executing your Commands via Motion Diffusion in Latent Space},
  author={Chen, Xin and  Jiang, Biao  Liu, Wen and Huang, Zilong and Fu, Bin and  Chen, Tao and  Yu, Gang},
  booktitle={Proceedings of the IEEE/CVF Conference on Computer Vision and Pattern Recognition (CVPR)},
  year={2023}
}

@inproceedings{temos,
  title     = {{TEMOS}: Generating diverse human motions from textual descriptions},
  author    = {Petrovich, Mathis and Black, Michael J. and Varol, G{\"u}l},
  booktitle = {European Conference on Computer Vision ({ECCV})},
  year      = {2022}
}

@inproceedings{ddpm,
author = {Ho, Jonathan and Jain, Ajay and Abbeel, Pieter},
title = {Denoising Diffusion Probabilistic Models},
year = {2020},
isbn = {9781713829546},
publisher = {Curran Associates Inc.},
address = {Red Hook, NY, USA},
booktitle = {Proceedings of the 34th International Conference on Neural Information Processing Systems (NeurIPS)},
articleno = {574},
location = {Vancouver, BC, Canada},
series = {NIPS'20}
}

@InProceedings{MAE,
    author    = {He, Kaiming and Chen, Xinlei and Xie, Saining and Li, Yanghao and Doll\'ar, Piotr and Girshick, Ross},
    title     = {Masked Autoencoders Are Scalable Vision Learners},
    booktitle = {Proceedings of the IEEE/CVF Conference on Computer Vision and Pattern Recognition (CVPR)},
    month     = {June},
    year      = {2022},
}

@inproceedings{CLIP,
  title={Learning Transferable Visual Models From Natural Language Supervision},
  author={Luo, Zhenzhong and Xiong, Yuwen and Sun, Xuancheng and Long, Yang and Yao, Ting and Mei, Tao},
  booktitle={Proceedings of the IEEE/CVF Conference on Computer Vision and Pattern Recognition (CVPR)},
  year={2020}
}

@inproceedings{
ddim,
title={Denoising Diffusion Implicit Models},
author={Jiaming Song and Chenlin Meng and Stefano Ermon},
booktitle={International Conference on Learning Representations (ICLR)},
year={2021}
}

@InProceedings{in2in,
    author    = {Ruiz-Ponce, Pablo and Barquero, German and Palmero, Cristina and Escalera, Sergio and Garc{\'\i}a-Rodr{\'\i}guez, Jos\'e},
    title     = {in2IN: Leveraging Individual Information to Generate Human INteractions},
    booktitle = {Proceedings of the IEEE/CVF Conference on Computer Vision and Pattern Recognition (CVPR) Workshops},
    month     = {June},
    year      = {2024},
    pages     = {1941-1951}
}

@article{vaswani2017attention,
  title={Attention is all you need},
  author={Vaswani, Ashish and Shazeer, Noam and Parmar, Niki and Uszkoreit, Jakob and Jones, Llion and Gomez, Aidan N and Kaiser, {\L}ukasz and Polosukhin, Illia},
  journal={Advances in neural information processing systems (NeurIPS)},
  volume={30},
  year={2017}
}

@inproceedings{fgt2m,
  title={Fg-t2m: Fine-grained text-driven human motion generation via diffusion model},
  author={Wang, Yin and Leng, Zhiying and Li, Frederick WB and Wu, Shun-Cheng and Liang, Xiaohui},
  booktitle={Proceedings of the IEEE/CVF International Conference on Computer Vision (ICCV)},
  pages={22035--22044},
  year={2023}
}

@inproceedings{attt2m,
  title={Attt2m: Text-driven human motion generation with multi-perspective attention mechanism},
  author={Zhong, Chongyang and Hu, Lei and Zhang, Zihao and Xia, Shihong},
  booktitle={Proceedings of the IEEE/CVF International Conference on Computer Vision (ICCV)},
  pages={509--519},
  year={2023}
}

@article{motiondiffuse,
  title={Motiondiffuse: Text-driven human motion generation with diffusion model},
  author={Zhang, Mingyuan and Cai, Zhongang and Pan, Liang and Hong, Fangzhou and Guo, Xinying and Yang, Lei and Liu, Ziwei},
  journal={IEEE Transactions on Pattern Analysis and Machine Intelligence (PAMI)},
  year={2024},
  publisher={IEEE}
}

@inproceedings{dit,
  author    = {Peebles, William and Xie, Saining},
  title     = {Scalable Diffusion Models with Transformers},
  booktitle = {Proceedings of the IEEE/CVF International Conference on Computer Vision (ICCV)},
  pages     = {4172--4182},
  year      = {2023},
  publisher = {IEEE Computer Society},
  month     = {October}
}

@inproceedings{ardiff,
  author    = {Li, Tao and Tian, Yu and Li, Hang and Deng, Mingyuan and He, Kaiming},
  title     = {Autoregressive Image Generation without Vector Quantization},
  booktitle = {Proceedings of the 38th Annual Conference on Neural Information Processing Systems (NeurIPS)},
  year      = {2024},
  note      = {To appear}
}

@inproceedings{armd,
  title={AMD: Autoregressive Motion Diffusion},
  author={Han, Bo and Peng, Hao and Dong, Minjing and Ren, Yi and Shen, Yixuan and Xu, Chang},
  booktitle={Proceedings of the AAAI Conference on Artificial Intelligence (AAAI)},
  volume={38},
  pages={2022--2030},
  year={2024}
}

@inproceedings{aamdm,
  title={AAMDM: Accelerated Auto-regressive Motion Diffusion Model},
  author={Li, Tianyu and Qiao, Calvin and Ren, Guanqiao and Yin, KangKang and Ha, Sehoon},
  booktitle={Proceedings of the IEEE/CVF Conference on Computer Vision and Pattern Recognition (CVPR)},
  pages={1813--1823},
  year={2024}
}

@inproceedings{
   interdiff,
   title={{InterDiff}: Generating 3D Human-Object Interactions with Physics-Informed Diffusion},
   author={Xu, Sirui and Li, Zhengyuan and Wang, Yu-Xiong and Gui, Liang-Yan},
   booktitle={ICCV},
   year={2023},
}

@inproceedings{cfg,
  title={Classifier-Free Diffusion Guidance},
  author={Ho, Jonathan and Salimans, Tim},
  booktitle={NeurIPS 2021 Workshop on Deep Generative Models and Downstream Applications},
  year={2021}
}

@inproceedings{guo2024momask,
  title={Momask: Generative masked modeling of 3d human motions},
  author={Guo, Chuan and Mu, Yuxuan and Javed, Muhammad Gohar and Wang, Sen and Cheng, Li},
  booktitle={Proceedings of the IEEE/CVF Conference on Computer Vision and Pattern Recognition (CVPR)},
  pages={1900--1910},
  year={2024}
}

@article{physhoi,
  title={Physhoi: Physics-based imitation of dynamic human-object interaction},
  author={Wang, Yinhuai and Lin, Jing and Zeng, Ailing and Luo, Zhengyi and Zhang, Jian and Zhang, Lei},
  journal={arXiv preprint arXiv:2312.04393},
  year={2023}
}

@inproceedings{tian2024var,
  author    = {Tian, Keyu and Jiang, Yicheng and Yuan, Ziyang and Peng, Bo and Wang, Lijuan},
  title     = {Visual Autoregressive Modeling: Scalable Image Generation via Next-Scale Prediction},
  booktitle = {Proceedings of the 37th Annual Conference on Neural Information Processing Systems (NeurIPS)},
  year      = {2024},
  pages     = {84839--84865},
  note      = {NeurIPS 2024 Best Paper}
}

@inproceedings{lfqvae,title	= {Language Model Beats Diffusion -- Tokenizer is Key to Visual Generation},author	= {Lijun Yu and José Lezama and Nitesh Bharadwaj Gundavarapu and Luca Versari and Kihyuk Sohn and David Minnen and Yong Cheng and Agrim Gupta and Xiuye Gu and Alex Hauptmann and Boqing Gong and Ming-Hsuan Yang and Irfan Essa and David Ross and Lu Jiang},year	= {2024},URL	= {https://openreview.net/forum?id=gzqrANCF4g},booktitle	= {ICLR}}

@inproceedings{ardhoi,
  title={Auto-regressive diffusion for generating 3d human-object interactions},
  author={Geng, Zichen and Hayder, Zeeshan and Liu, Wei and Mian, Ajmal Saeed},
  booktitle={Proceedings of the AAAI Conference on Artificial Intelligence (AAAI)},
  volume={39},
  number={3},
  pages={3131--3139},
  year={2025}
}

@inproceedings{wang2025timotion,
  title={TIMotion: Temporal and Interactive Framework for Efficient Human-Human Motion Generation},
  author={Wang, Yabiao and Wang, Shuo and Zhang, Jiangning and Fan, Ke and Wu, Jiafu and Xue, Zhucun and Liu, Yong},
  booktitle={Proceedings of the Computer Vision and Pattern Recognition Conference (CVPR)},
  pages={7169--7178},
  year={2025}
}

@inproceedings{xu2024regennet,
  title={ReGenNet: Towards Human Action-Reaction Synthesis},
  author={Xu, Liang and Zhou, Yizhou and Yan, Yichao and Jin, Xin and Zhu, Wenhan and Rao, Fengyun and Yang, Xiaokang and Zeng, Wenjun},
  booktitle={CVPR},
  pages={1759--1769},
  year={2024}
}

@article{ji2025humanx,
  title={Towards Immersive Human-X Interaction: A Real-Time Framework for Physically Plausible Motion Synthesis},
  author={Ji, Kaiyang and Shi, Ye and Jin, Zichen and Chen, Kangyi and Xu, Lan and Ma, Yuexin and Yu, Jingyi and Wang, Jingya},
  journal={arXiv preprint arXiv:2508.02106},
  year={2025}
}

@article{geng2025mean,
  title={Mean Flows for One-step Generative Modeling},
  author={Geng, Zhengyang and Deng, Mingyang and Bai, Xingjian and Kolter, J Zico and He, Kaiming},
  journal={arXiv preprint arXiv:2505.13447},
  year={2025}
}

@inproceedings{flowmatching,
  title={Flow Matching for Generative Modeling},
  author={Lipman, Yaron and Chen, Ricky TQ and Ben-Hamu, Heli and Nickel, Maximilian and Le, Matt},
  booktitle={11th International Conference on Learning Representations (ICLR)},
  year={2023}
}

@inproceedings{rectifiedflow,
  title={Flow Straight and Fast: Learning to Generate and Transfer Data with Rectified Flow},
  author={Liu, Xingchao and Gong, Chengyue and others},
  booktitle={The Eleventh International Conference on Learning Representations (ICLR)},
  year={2023}
}

@article{ntu_rgbd120,
  title={NTU RGB+D 120: A large-scale benchmark for 3D human activity understanding},
  author={Liu, Jun and Shahroudy, Amir and Perez, Mauricio and Wang, Gang and Duan, Ling-Yu and Kot, Alex C},
  journal={IEEE Transactions on Pattern Analysis and Machine Intelligence (PAMI)},
  volume={42},
  number={10},
  pages={2684--2701},
  year={2020}
}

@InProceedings{chi3d,
author = {Fieraru, Mihai and Zanfir, Mihai and Oneata, Elisabeta and Popa, Alin-Ionut and Olaru, Vlad and Sminchisescu, Cristian},
title = {Three-Dimensional Reconstruction of Human Interactions},
booktitle = {The IEEE/CVF Conference on Computer Vision and Pattern Recognition (CVPR)},
month = {June},
year = {2020}
}

@inproceedings{camdm,
  title={Taming Diffusion Probabilistic Models for Character Control},
  author = {Chen, Rui and Shi, Mingyi and Huang, Shaoli and Tan, Ping and Komura, Taku and Chen, Xuelin},
  year = {2024},
  publisher = {Association for Computing Machinery},
  address = {New York, NY, USA},
  url = {https://doi.org/10.1145/3641519.3657440},
  doi = {10.1145/3641519.3657440},
  booktitle = {ACM SIGGRAPH 2024 Conference Papers},
  location = {Denver, CO, USA},
  series = {SIGGRAPH '24}
}

@inproceedings{tokenflow,
  title={TokenFlow: Consistent Diffusion Features for Consistent Video Editing},
  author={Geyer, Michal and Bar-Tal, Omer and Bagon, Shai and Dekel, Tali},
  booktitle={The Twelfth International Conference on Learning Representations (ICLR)},
  year={2024}
}

@article{nextstep,
  title={NextStep-1: Toward Autoregressive Image Generation with Continuous Tokens at Scale},
  author={NextStep Team and Chunrui Han and Guopeng Li and Jingwei Wu and Quan Sun and Yan Cai and Yuang Peng and Zheng Ge and Deyu Zhou and Haomiao Tang and Hongyu Zhou and Kenkun Liu and Ailin Huang and Bin Wang and Changxin Miao and Deshan Sun and En Yu and Fukun Yin and Gang Yu and Hao Nie and Haoran Lv and Hanpeng Hu and Jia Wang and Jian Zhou and Jianjian Sun and Kaijun Tan and Kang An and Kangheng Lin and Liang Zhao and Mei Chen and Peng Xing and Rui Wang and Shiyu Liu and Shutao Xia and Tianhao You and Wei Ji and Xianfang Zeng and Xin Han and Xuelin Zhang and Yana Wei and Yanming Xu and Yimin Jiang and Yingming Wang and Yu Zhou and Yucheng Han and Ziyang Meng and Binxing Jiao and Daxin Jiang and Xiangyu Zhang and Yibo Zhu},
  journal={arXiv preprint arXiv:2508.10711},
  year={2025}
}

@article{wang2025marrs,
  title={MARRS: Masked Autoregressive Unit-based Reaction Synthesis},
  author={Wang, YB and Wang, Shuo and Zhang, JN and Wu, JF and He, QD and Fu, CC and Wang, CJ and Liu, Yong},
  journal={arXiv preprint arXiv:2505.11334},
  year={2025}
}

@ARTICLE{interformer,
  author={Chopin, Baptiste and Tang, Hao and Otberdout, Naima and Daoudi, Mohamed and Sebe, Nicu},
  journal={IEEE Transactions on Multimedia (TMM)}, 
  title={Interaction Transformer for Human Reaction Generation}, 
  year={2023},
  volume={},
  number={},
  pages={1-13},
  doi={10.1109/TMM.2023.3242152}}

@InProceedings{maskgit,
  title = {MaskGIT: Masked Generative Image Transformer},
  author={Huiwen Chang and Han Zhang and Lu Jiang and Ce Liu and William T. Freeman},
  booktitle = {The IEEE Conference on Computer Vision and Pattern Recognition (CVPR)},
  month = {June},
  year = {2022}
}

@inproceedings{ARDM,
  title={Autoregressive Diffusion Models},
  author={Hoogeboom, Emiel and Gritsenko, Alexey A and Bastings, Jasmijn and Poole, Ben and van den Berg, Rianne and Salimans, Tim},
  booktitle={International Conference on Learning Representations (ICLR)},
  year={2022}
}

@inproceedings{cen2025ready_to_react,
  title={Ready-to-React: Online Reaction Policy for Two-Character Interaction Generation},
  author={Cen, Zhi and Pi, Huaijin and Peng, Sida and Shuai, Qing and Shen, Yujun and Bao, Hujun and Zhou, Xiaowei and Hu, Ruizhen},
  booktitle={ICLR},
  year={2025}
}

@article{Diff9D,
  author={Liu, Jian and Sun, Wei and Yang, Hui and Deng, Pengchao and Liu, Chongpei and Sebe, Nicu and Rahmani, Hossein and Mian, Ajmal},
  journal={IEEE Transactions on Pattern Analysis and Machine Intelligence},
  title={Diff9D: Diffusion-Based Domain-Generalized Category-Level 9-DoF Object Pose Estimation},
  year={2025},
  volume={47},
  number={7},
  pages={5520-5537},
  doi={10.1109/TPAMI.2025.3552132}
}
}


\end{document}